\documentclass{article}
\usepackage{amssymb}

\usepackage[final]{corl_2025} % Uncomment for the camera-ready ``final'' version.

\usepackage{hyperref}
\usepackage{url}
\usepackage{graphicx}
\usepackage{subfigure}
\usepackage{wrapfig}
\usepackage{multirow}
\usepackage{array}
\usepackage{amsmath}
\usepackage{amssymb}
\usepackage{mathtools}
\usepackage{amsthm}
\usepackage{algorithmic}
\usepackage{algorithm}
\usepackage{booktabs} 
\usepackage{amsmath,amsfonts,bm}

\def\eqref#1{equation~\ref{#1}}
\def\1{\bm{1}}

\DeclareMathAlphabet{\mathsfit}{\encodingdefault}{\sfdefault}{m}{sl}
\SetMathAlphabet{\mathsfit}{bold}{\encodingdefault}{\sfdefault}{bx}{n}

\DeclareMathOperator*{\argmax}{arg\,max}

\newcommand{\methodabbr}{SDAX}

\title{Unsupervised Skill Discovery as Exploration \\ for Learning Agile Locomotion}
\author{
  Seungeun Rho\thanks{Co-first authors. Project page - \href{seungeunrho.github.io/projects/SDAX/}{https://seungeunrho.github.io/projects/SDAX/}} , Kartik Garg\footnotemark[1] , Morgan Byrd, Sehoon Ha\\
  Georgia Institute of Technology\\
  \texttt{\{srho31, kgarg65, abyrd45, sehoonha\}@gatech.edu} \\
}

\begin{document}
\maketitle

%===============================================================================

\begin{abstract}
Exploration is crucial for enabling legged robots to learn agile locomotion behaviors that can overcome diverse obstacles. However, such exploration is inherently challenging, and we often rely on extensive reward engineering, expert demonstrations, or curriculum learning—all of which limit generalizability.
In this work, we propose Skill Discovery as Exploration (\textbf{\methodabbr}), a novel learning framework that significantly reduces human engineering effort. \methodabbr\ leverages unsupervised skill discovery to autonomously acquire a diverse repertoire of skills for overcoming obstacles. To dynamically regulate the level of exploration during training, \methodabbr\ employs a bi-level optimization process that autonomously adjusts the degree of exploration.
We demonstrate that \methodabbr\ enables quadrupedal robots to acquire highly agile behaviors—including crawling, climbing, leaping, and executing complex maneuvers such as jumping off vertical walls. Finally, we deploy the learned policy on real hardware, validating its successful transfer to the real world.

\end{abstract}


%===============================================================================
\begin{figure}[h]
    \centering
    \includegraphics[height=1.3in]{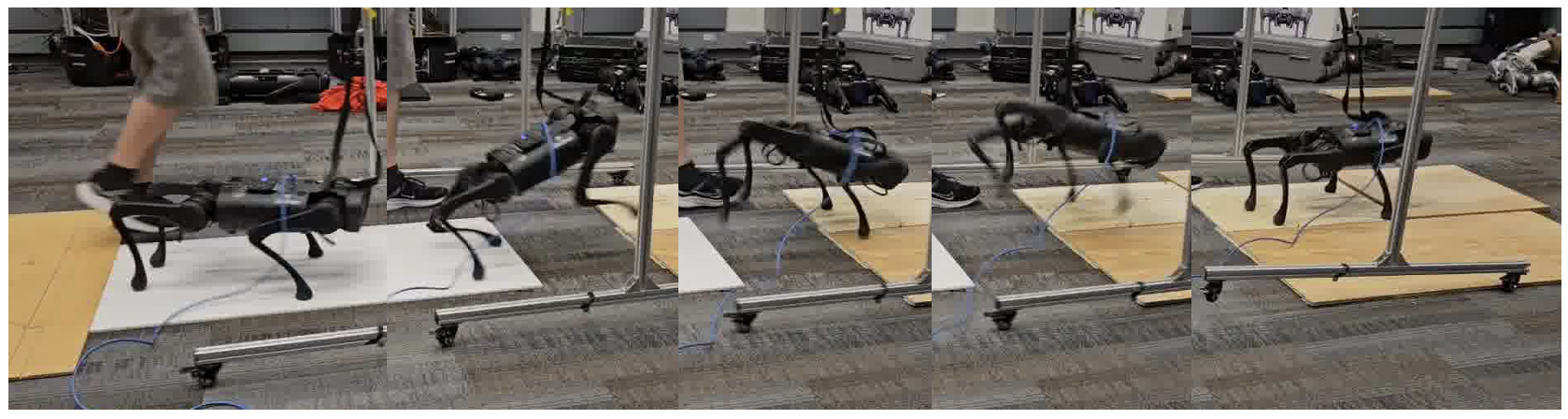}
    \DeclareGraphicsExtensions.
\caption{We deployed our policy on a real robot. The robot successfully leaps over the gap.}
\label{fig_hardware_leap}
\end{figure}

\section{Introduction}

In recent years, combining legged robot locomotion with deep reinforcement learning (deep RL) has led to remarkable advances in agility~\citep{ref2, ref3, ref4, ref5, ref6, ref7, ref8, luo2024pie}. However, existing methods often depend on additional techniques to master challenging skills, including: (1) reward engineering informed by domain expertise~\citep{robot-parkour, extreme-parkour, ref9}, (2) demonstration datasets~\citep{bogdanovic2022model, kilinc2022reinforcement, li2023learning, he2024learning}, and (3) carefully crafted curriculum learning~\citep{RMA}. In this work, we introduce \textbf{S}kill \textbf{D}iscovery \textbf{A}s e\textbf{X}ploration \textbf{(\methodabbr)}, a framework capable of solving highly challenging tasks—such as the \textit{wall-jump} shown in Figure~\ref{fig_walljump_solved}—by autonomously exploring a diverse set of strategies. \methodabbr\ leverages unsupervised reinforcement learning to achieve this, eliminating the need for manually designed curricula or demonstration data.

Unsupervised RL offers a framework for learning diverse \textit{skills}, where each skill corresponds to a highly correlated sequence of behaviors and is represented by a vector $z \in \mathbb{R}^n$. The intrinsic reward in unsupervised RL encourages the policy to exhibit different behaviors when conditioned on different skill vectors $z$, a process commonly referred to as \textit{unsupervised skill discovery}. We harness this learning process to explore a wide range of high-level action strategies, ultimately enabling the emergence of agile behaviors necessary to solve challenging tasks.

In detail, \methodabbr\ combines two objectives: \textbf{solving the given task} and \textbf{finding diverse solutions}. Solving the task is represented by maximizing the task reward. The task reward is kept simple, such as following forward velocity commands to move toward task completion. On the other hand, exploring diverse behaviors is achieved by maximizing a diversity reward, which is derived from skill discovery methods. This encourages the agent to try various approaches to find the desired height, orientation, velocity, or angular velocity needed to solve the task. However, balancing two distinct objectives is not straightfoward and one may overpower the other. If the task reward dominates, agents may not sufficiently explore diverse behaviors. Conversely, if the diversity reward dominates, agents may spend too much time exploring, failing to solve the task. This is analogous to the exploration-exploitation trade-off in RL \citep{sutton2018reinforcement}. To address this problem, we introduce a learnable parameter $\lambda$ to balance the two objectives. We train $\lambda$ to automatically adjust the weight of the diversity reward to maximize the task reward. 

In summary, \methodabbr\ aims to adopt skill discovery methods for high-level explorations to optimize the task-specific reward. The primary contributions of this work are as follows: (1) We propose a novel framework that combines RL and unsupervised skill discovery algorithms to automatically learn agile locomotion skills.
(2) We provide a thorough derivation of a bi-level optimization framework for training the balancing parameter $\lambda$. We also demonstrate that \methodabbr\ of adapting $\lambda$ robustly finds the optimal value for a given task. (3) We evaluate \methodabbr\ against manual exploration strategies on four challenging locomotion tasks: jumping, leaping, crawling, and a wall-jump. 

%===============================================================================
\section{Related Work}
\label{sec:related-work}

\paragraph{Unsupervised Skill Discovery.}
The goal of unsupervised skill discovery is to establish an association between a latent skill vector $z$ and the resulting skill-conditioned policy $\pi(a|s,z)$. Two main families of approaches have been proposed for achieving this goal. The first family maximizes the mutual information between skills and states, \( I(z; s) \). Examples include DIAYN~\citep{diayn} and VIC~\citep{gregor2016variational}. The second family maximizes the Wasserstein Dependency Measure, $I_{WDM}$, as seen in methods such as LSD~\citep{lsd}, METRA~\citep{metra}, and LGSD~\citep{rho2024language}. Since \methodabbr\ is agnostic to the choice of skill discovery approach, we evaluate it using representative algorithms from both families: DIAYN and METRA.

\paragraph{Learning Agile Locomotion.} Recently, learning-based methods have demonstrated highly agile locomotion capabilities such as high-speed running ~\citep{ref16, ref35}, jumping ~\citep{ref59, ref60}, and climbing ~\citep{ref62, ref10'}. Our work aims to cover not only jumping, running, and leaping, but also \textit{wall-jumping}, which involves a parkour-style motion combining flipping and jumping using walls.

The work most related to ours is that of \citet{robot-parkour}, which used a manually designed reward that penalizes the overlap between the robot and imaginary obstacles. They trained agents to minimize these overlaps, resulting in the learning of agile behaviors. In contrast, we aim to train a similar set of tasks without the need for such reward designs. Instead, we allow an unsupervised RL method to discover the skills required to solve these tasks.

\paragraph{Skill Discovery for Locomotion.} 

Recently, several approaches have incorporated skill discovery into locomotion training pipelines. \citet{cheng2024learning} used skill discovery methods to obtain diverse skills for solving tasks. A core difference is that they assume access to a near-optimal policy and value function from the start, and introduce skill discovery to diversify skills based on the given policy. In contrast, \methodabbr\ uses skill discovery in-the-loop to obtain an optimal policy for solving the task. \citet{atanassov2024constrained} proposed replacing the objective of LSD with a ``norm-matching'' objective to obtain more diverse and controllable skills. Our aim is not to replace or outperform existing skill discovery methods, but rather to leverage them as a high-level exploration strategy to acquire agile behaviors.

%===============================================================================
\section{Unsupervised Skill Discovery as Exploration}
\label{sec:methods}
\subsection{Problem Formulation}
We regard the problem of training a control module for a legged robot as a Markov Decision Process (MDP) defined as $\mathcal{M} \equiv \{\mathcal{S},\mathcal{A},\mathcal{R},\mathcal{P},\gamma \}$, where $\mathcal{S}$ is a state space, $\mathcal{A}$ is an action space composed of joint position targets for a PD controller of the robot, $\mathcal{R}$ is a reward function, $\mathcal{P}$ is a transition probability, and $\gamma$ is a discount factor. The objective of RL is to obtain an optimal policy $\pi$ which maximizes the expected sum of the discounted reward $J = \mathbb{E_{\pi}}\Big[\sum_{t=0}^{\infty} \gamma^t r_t\Big]$. $\pi$ can be parameterized with the neural network $\theta$, so here we denote policy as $\pi_\theta$. However, instead of training a standard policy $\pi_\theta(a|s)$, we train a skill-conditioned policy $\pi_\theta(a|s,z)$, where $z$ is randomly sampled from a fixed prior distribution, $z \sim p(z)$, for each episode and remains fixed throughout the episode. 

\subsection{Algorithm}
\label{our approach}
Our objective is to find the policy parameter $\theta$ that optimizes the expected sum of both the task reward $r^{\text{task}}$ and the diversity reward $r^{\text{div}}$.
$$
\theta = 
\argmax_{\theta}J^{\text{task+div}} = 
\argmax_{\theta}\mathbb{E_{\pi_{\theta}}}\Big[\sum_{t=0}^{\infty} \gamma^t (r^{\text{task}}_t + \lambda r^{\text{div}}_t)\Big] $$

A learnable parameter $\lambda$ determines the weight of $r^{\text{div}}$, and we refer to it as the balancing parameter. The task reward $r^{\text{task}}_t$ specifies the goal of the task. It can be defined for each task and should be kept simple, such as a forward velocity tracking reward. Regardless of the value of $\lambda$, the policy $\pi$ is always conditioned on a particular $z$. Conditioning the policy on different values of $z$ results in different behaviors, so training a skill-conditioned policy with $\lambda=0$ effectively means we are training a group of different policies, all of which converge into a single behavior. When $\lambda$ becomes large, the diversity reward dominates, and each policy learns a distinct skill, but none of them are capable of solving the task. Thus, determining the appropriate value of $\lambda$ is crucial. In the following paragraphs, we will explain how the balancing parameter $\lambda$ is trained and how $r^{\text{div}}$ is defined.

\begin{figure}[t]
    \centering
    \includegraphics[height=1.3in]{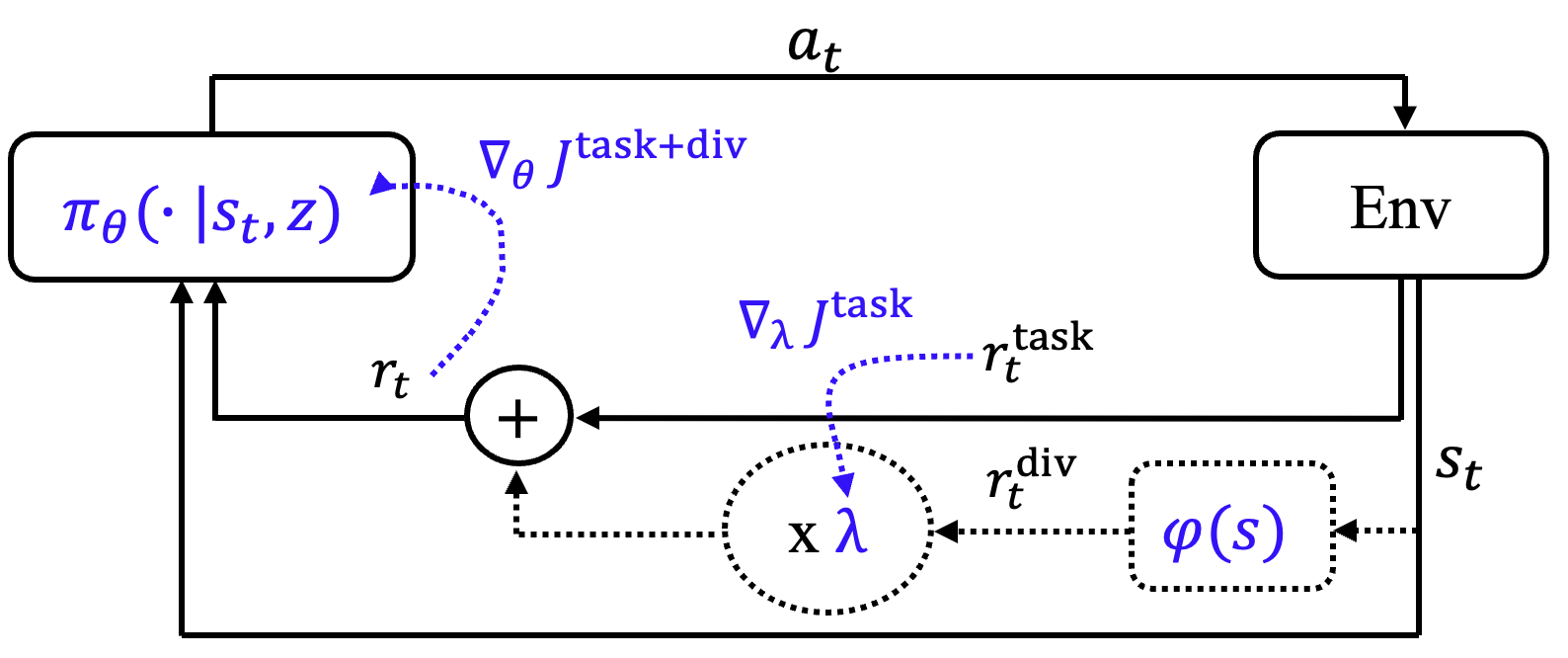}
    \DeclareGraphicsExtensions.
\caption{A figure of bi-level optimization for $\pi_{\theta}$ and $\lambda$. The task reward gives the gradient signal for training $\lambda$, and the sum of both sources of rewards provides the gradient signal for optimizing $\pi_{\theta}$. 
}
\label{fig_method_overall}
\end{figure}

\paragraph{Train Balancing Parameter}
\label{paragraph_train_balance_parameter}

As depicted in the Figure~\ref{fig_method_overall}, we utilize a bi-level optimization framework to train both policy $\pi$ and a learnable balancing parameter $\lambda$, which is similar to LIRPG~\citep{zheng2018learning}.
While $\theta$ is trained to maximize $J^{\text{task+div}}$, $\lambda$ is trained to maximize only $J^{\text{task}}=\mathbb{E_{\pi_{\theta}}}\Big[\sum_{t=0}^{\infty} \gamma^t r^{\text{task}}_t\Big]$. It is worth noting that our ultimate goal is to solve the given task. So, the intuitive meaning of training $\lambda$ solely depending on the task reward is that we determine the degree of diversity reward only to maximize the task performance. Ideally, when the diversity reward helps solve the task, $\lambda$ will be increased, and if it rather deters training, $\lambda$ will be decreased. 

More concretely,
\begin{equation}
\label{chain_rule_decomposition}
\lambda = \argmax_{\lambda} J^{\text{task}}.
\end{equation}
The challenge here is that we cannot directly compute the gradient of $J^{\text{task}}$ with respect to $\lambda$. To address this, we apply the chain rule to decompose the gradient as follows: \begin{equation} \nabla_{\lambda}J^{\text{task}} = \nabla_{\theta}J^{\text{task}} \nabla_{\lambda}{\theta}. \end{equation}

To make this expression tractable, it can be further expanded into the following final form: \begin{equation} \label{equation_last} \nabla_{\lambda}J^{\text{task}} \approx  \alpha A^{\text{task}} \nabla_{\theta'} \log \pi_{\theta'}(a | s, z) \cdot  A^{\text{div}} \nabla_{\theta} \log \pi_{\theta}(a | s, z).\end{equation}

Here, $\alpha$ is the learning rate, $\theta'$ denotes the parameter $\theta$ after a single update, and $A^{\text{task}}$ and $A^{\text{div}}$ denote the advantage values computed using $r^{\text{task}}$ and $r^{\text{div}}$, respectively. The detailed derivation of Equation (\ref{equation_last}) is provided in Appendix~\ref{proof_of_bilevel_optimization}. Now we can directly compute this term using a sample-based approximation. 

The intuitive meaning of this formula is that if the gradient vectors from the task reward and the diversity reward point in a similar direction, $\lambda$ should be increased; otherwise, it should be decreased.  The key difference between \methodabbr\ and ~\citet{zheng2018learning} is that instead of training the intrinsic reward function itself, we fix the intrinsic reward as the diversity reward, and we only train the balancing parameter $\lambda$ to determine the degree of it.

\paragraph{Diversity Reward}
For the diversity reward $r^{\text{div}}$, we follow the formulation of METRA~\citep{metra}. They train skills to maximize the Wasserstein Dependency Measure~\citep{ozair2019wasserstein} $I_{\text{WDM}} = I_{W}(S;Z)$. Maximization of $I_{\text{WDM}}$ can be translated into the following objective:
$$
\sup_{\pi, \phi} \mathbb{E}_{P(\tau, z)} \left[ \sum_{t=0}^{T-1} (\phi(s_{t+1}) - \phi(s_{t}))^T z \right] \text{ s.t. } \| \phi(s) - \phi(s') \|_2 \leq 1, \forall (s, s') \in \mathcal{S}_{\text{adj}},
$$
Here, $\phi:S \rightarrow Z$ is a learnable representation function that maps the state into a latent skill space. Optimization of this term can be achieved by simply using an off-the-shelf RL algorithm to maximize the reward $r^{\text{div}} = (\phi(s_{t+1}) - \phi(s_{t}))^T z $. To ensure that $\phi$ satisfies the constraint, we use dual gradient descent
with a Lagrange multiplier $\kappa$ with a small margin $\epsilon > 0$. Please refer to \citet{metra} for more details. 

\paragraph{Skill Selection}
A typical unsupervised skill discovery method requires careful selection of the optimal skill vector $z$ during the testing phase. However, we observed that as training progresses, an increasing proportion of the learned skills exhibit successful behaviors, a phenomenon we refer to as ``positive collapse''(Section~\ref{sec:convergence}). Therefore, in this work, we simply select a random skill $z$ for reporting performance, rather than selectively choosing it or training a high-level controller.

\paragraph{Implementation Details}
We introduced two separate value networks, $v^{\text{task}}_{\psi_1}$ and $v^{\text{div}}_{\psi_2}$, due to the presence of two distinct reward sources: $r^{\text{task}}$ and $r^{\text{div}}$. Using a single value network to model the value of $r^{\text{task}} + \lambda r^{\text{div}}$ led to unstable training, as the scale of the rewards varied with changes in $\lambda$. Pseudo-code for our algorithm is provided in the appendix \ref{algorithm_ours}.

%===============================================================================
\section{Experimental Results}

In this section, we evaluate the proposed framework by training policies on a set of agile locomotion tasks. First, we examine three robot parkour learning tasks from \citet{robot-parkour}, including leaping, climbing, and crawling,  which require distinctive control strategies to overcome obstacles. On these tasks, we experiment with how skill discovery methods can aid in learning agile behaviors and evaluate \methodabbr\ against baselines. 

We use Isaac Gym~\citep{isaacgym} as the simulation engine, and our codebase builds on the work of \citet{rudin2022learning}. All experiments are conducted using the Unitree A1 robot. The gap for the leap task is 48cm, the platform height for the climb task is 25cm, and the gap height for the crawl task is 29cm. We adopt Proximal Policy Optimization (PPO)~\citep{ppo} as our main reinforcement learning algorithm, and details of the observation space are provided in Appendix~\ref{appendix_observation}. Depending on the task, policies typically converge within 10k–20k iterations, requiring approximately 8–16 hours of training on an NVIDIA A40 GPU.

\subsection{Learning Agile Locomotion Skills}

We compared \methodabbr\ against the following baseline algorithms:
\begin{itemize}
    \item \textit{Task-only}: An RL baseline trained only with task specific rewards $r^{\text{task}}$. 
    \item \textit{Div-only}: An RL baseline trained using diversity reward $r^{\text{div}}$ only. 
    \item \textit{RND}: It combines $r^{\text{task}}$ with an exploration reward instead of a diversity reward. 
    \item \textit{\methodabbr\ with DIAYN}: $r^{\text{div}}$ is computed with DIAYN reward. 
    \item \textit{\methodabbr\ with METRA}: $r^{\text{div}}$ is computed with METRA reward.
\end{itemize}
We designed the same task reward across all baseline methods and tasks, with the primary goal of incentivizing agents to move forward. Details of the task rewards are provided in Appendix \ref{appendix_reward}. Since \methodabbr\ can be regarded as an exploration mechanism, we included RND~\citep{burda2018exploration}, one of the most widely adopted exploration algorithms, as a baseline.  For both the diversity reward and exploration bonus in RND, we manually specify sub-dimensions of the state space, ensuring that the learning process focuses on exploration within the specified sub-dimensions. Specifically, we selected base heights for climbing and crawling tasks and forward velocity for leaping. Additionally, to expedite the learning of the \textit{Div-only} agent, we provided the robot's base $x$ position as an additional input to the skill discovery algorithm. This facilitated the exploration of diverse $x$ positions, ultimately helping the agent move forward.

\paragraph{\methodabbr\ enables learning the skills needed to solve each task.}

\begin{figure*}[t]
    \centering
    \subfigure[Leap]
    {\includegraphics[width=0.34\textwidth]{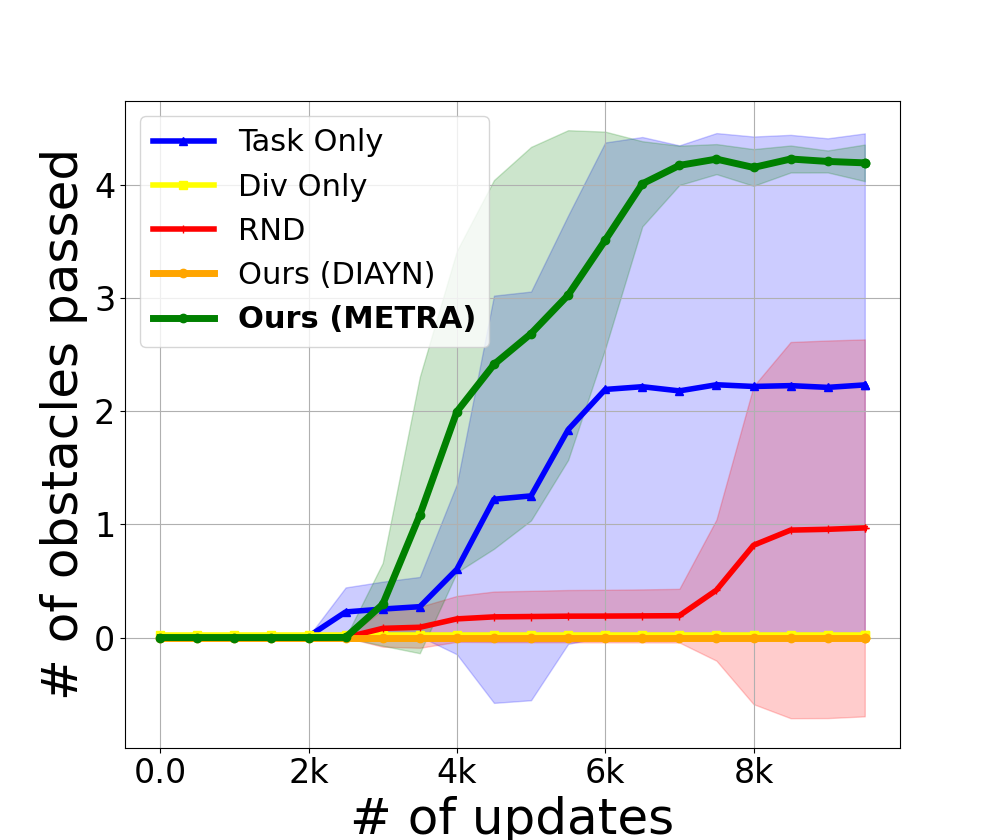}}
    \hspace*{-5mm}
    \subfigure[Climb]
    {\includegraphics[width=0.34\textwidth]{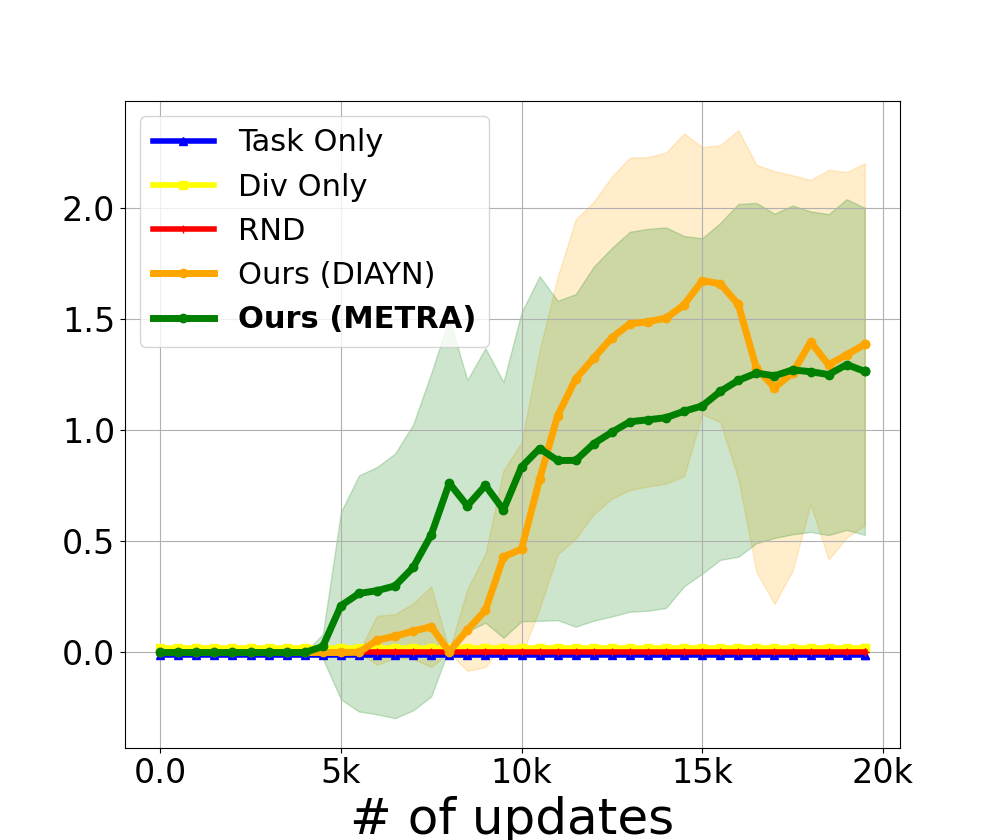}} 
    \hspace*{-5mm}
    \subfigure[Crawl]
    {\includegraphics[width=0.34\textwidth]{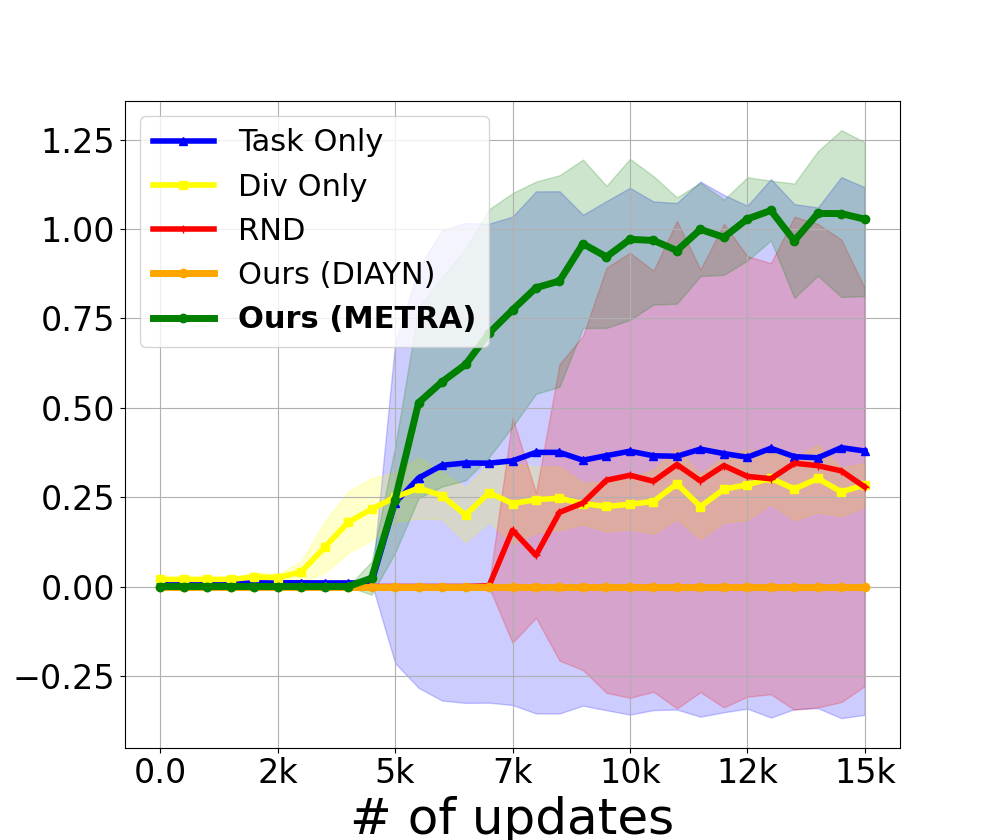}} 
    \caption{Training curves, which denote the number of objects passed over the number of updates. Our method with METRA can solve all the tasks and exhibits better sample efficiency.}
    \label{fig_parkour_learning_curve}
\end{figure*}

We present the training curves of \methodabbr\ and all baseline algorithms in Figure~\ref{fig_parkour_learning_curve} over five different seeds. We measured the number of obstacles passed in each task, where each task contains three consecutive obstacles of the same configuration. \methodabbr\ successfully learned the necessary motor skills for all tasks.
Compared to the \textit{Task-only} baseline, we observed that incorporating diversity rewards helps in learning agile locomotion skills. However, relying solely on diversity rewards (\textit{Div-only}) fails to achieve meaningful skills, highlighting that a balanced interplay between task and diversity rewards is critical for success. Additionally, a comparison with \textit{RND} shows that diversity-based approaches outperform exploration-based rewards. We believe this is because naive exploration-based methods focus on state-level exploration, incentivizing agents to visit nearby unvisited states, making skill-level exploration challenging. In contrast, skill discovery methods inherently facilitate skill-level exploration, as they encourage skills to explore distinct sets of states, allowing agents to transition to entirely new regions. Lastly, it is worth noting that METRA outperforms DIAYN as a skill discovery module. This is because the diversity reward $r^{\text{div}}$ from DIAYN can be maximized even with small differences between states, as long as the discriminator network can distinguish them, whereas METRA seeks diversity without saturation.

\paragraph{Skill discovery enables high level exploration.}

\begin{figure*}[t]
    \centering
    \subfigure[Leap - success]
    {\includegraphics[width=0.32\textwidth]{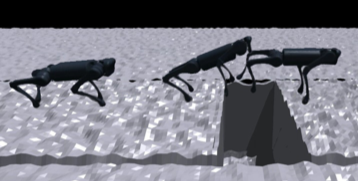}} 
    \vspace*{-1.0mm}
    \subfigure[Climb - success]
    {\includegraphics[width=0.32\textwidth]{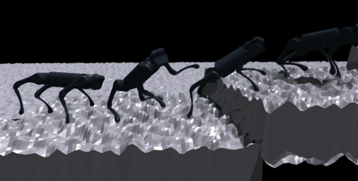}} 
    \vspace*{-1.0mm}
    \subfigure[Crawl - success]
    {\includegraphics[width=0.32\textwidth]{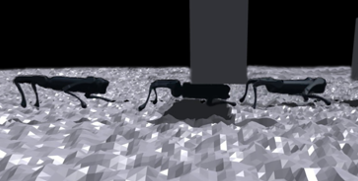}} 
    \vspace*{-1.0mm}
    \subfigure[Leap - failure]
    {\includegraphics[width=0.32\textwidth]{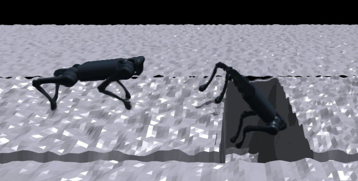}}
    % \hspace*{-2mm}
    \subfigure[Climb - failure]
    {\includegraphics[width=0.32\textwidth]{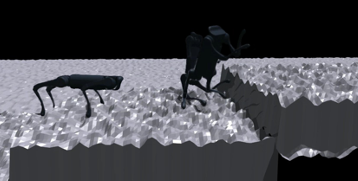}}
    % \hspace*{-2mm}
    \subfigure[Crawl - failure]
    {\includegraphics[width=0.32\textwidth]{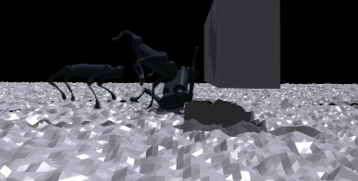}} 
    % \hspace*{-2mm}
    \caption{Visualization of the diverse skills explored by the robot during training.
    }
    \vspace*{-5.0mm}
    
    \label{fig_skill_comparion}
\end{figure*}

We also provide qualitative evidence demonstrating how skill discovery methods enhance exploration. Figure~\ref{fig_skill_comparion} illustrates example behaviors of \methodabbr\ using two different skills for each task based on an actual model checkpoint from training. To observe the behaviors of different skills, we kept the model fixed and fed different skill vectors to the policy. As a result, both successful and unsuccessful episodes were generated from the same policy, using different skill vectors. 
In the leaping task, some skills enabled the agent to powerfully kick off the ground, gaining enough height to clear the gap, while others resulted in weak jumps that led to failure. In the crawling task, certain skills lowered the robot’s body posture to pass under the obstacle effectively, whereas others caused the robot to jump and lose balance. These examples illustrate that the agent explores a diverse range of heights—some of which solve the task while others do not. When a particular skill starts solving the task, the task reward increases, leading to successful task completion. In this sense, skill discovery functions as a high-level exploration module.

\subsection{Learning Balancing Parameter $\lambda$}

\begin{figure*}[t]
    
    \centering
    \subfigure[Training curve]
    {\includegraphics[width=0.33\textwidth]
    {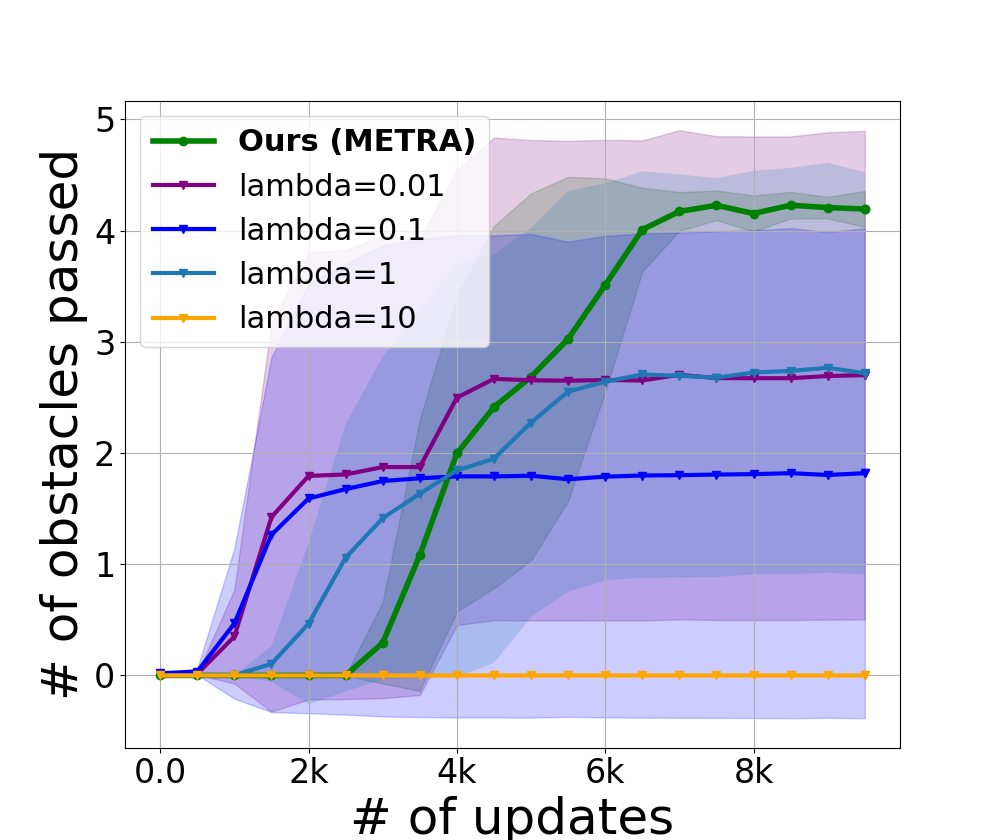}} 
    \subfigure[Corresponding curve of $\lambda$]
    {\includegraphics[width=0.33\textwidth]{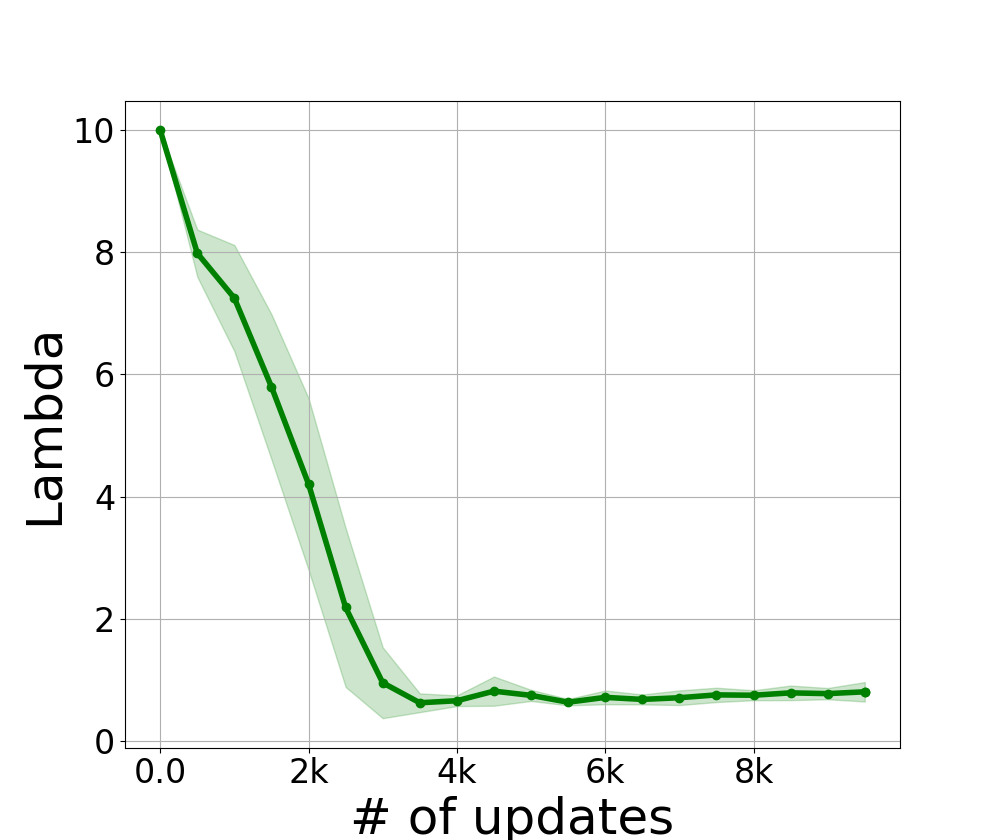}} 
    \caption{\methodabbr\ outperforms all the baseline rewards with fixed value of lambda.}
    \vspace*{-5.0mm}
    \label{fig_vs_fixed_kappa}
\end{figure*}

Selecting the appropriate value for $\lambda$ is crucial, as the scale of both the task reward and diversity reward is difficult to determine a priori. If either the task reward or the diversity reward dominates, the agent's learning process can be significantly hindered. In this section, we demonstrate how our algorithm effectively adjusts $\lambda$ during training. We compare our adaptive approach to fixed values of $\lambda$, using four different settings: ${0.01, 0.1, 1, 10}$. These experiments were conducted on the leaping tasks from the previous section, with each method trained using three different random seeds. We measured performance based on the number of obstacles passed.

\paragraph{Our method outperforms fixed $\lambda$ values.}

Figure~\ref{fig_vs_fixed_kappa}(a) shows that our adaptive method outperforms all fixed-value experiments. \methodabbr\ demonstrated both superior sample efficiency and final performance compared to rest of the $\lambda$ values. Figure \ref{fig_vs_fixed_kappa}(b) illustrates how the learned $\lambda$ values evolve during training. The value starts at 10.0 and gradually decreases, suggesting that our algorithm learned that decreasing $\lambda$ helps maximize task rewards over time. 

It is also important to note that our method does not correspond to a single fixed $\lambda$ value throughout training. In other words, there may not exist a single value of $\lambda$ that could yield an identical training curve. \methodabbr\ adjusts $\lambda$ dynamically, resulting in different values at different stages of training, which allows the agent to achieve an appropriate balance of diversity and task reward throughout the learning process.

\subsection{Convergence of Different Skills into a Narrow Solution Space}
\label{sec:convergence}

% \begin{table}[t]
%     \centering
%     \begin{tabular}{p{1.15cm}p{1.15cm}p{1.15cm}|p{1.15cm}p{1.15cm}p{1.15cm}|p{1.15cm}p{1.15cm}p{1.15cm}}
%         \toprule
%         \multicolumn{3}{p{3.45cm}|}{\textbf{Leap}} & 
%         \multicolumn{3}{p{3.45cm}|}{\textbf{Climb}} & 
%         \multicolumn{3}{p{3.45cm}}{\textbf{Crawl}} \\ 
        
%         1k & 2k & 3k & 12k & 15k & 20k & 2k & 7k & 15k \\ 
%         \midrule
        
%         29.9$\pm$5.2 & 99.1$\pm$0.9 &99.4$\pm$0.8 & 
        
%         49.4$\pm$7.2 & 71$\pm$9.3 &68.7$\pm$11 & 
        
%         22.3$\pm$3.1 & 31.7$\pm$3.8 &40$\pm$2.8  \\ \bottomrule
%         % Add more rows as needed
%     \end{tabular}
\begin{table}[t]
    \centering
    \begin{tabular}{p{1.15cm}p{1.15cm}p{1.15cm}|p{1.15cm}p{1.15cm}p{1.15cm}|p{1.15cm}p{1.15cm}p{1.15cm}}
        \toprule
        \multicolumn{3}{p{3.45cm}|}{\textbf{Leap}} & 
        \multicolumn{3}{p{3.45cm}|}{\textbf{Climb}} & 
        \multicolumn{3}{p{3.45cm}}{\textbf{Crawl}} \\ 
        
        3k & 5k & 10k & 7k & 10k & 20k & 5k & 10k & 15k \\ 
        \midrule
        
        43.1$\pm$4.3 & 90.6$\pm$2.6 &97.1$\pm$2.3 & 
        
        31.0$\pm$3.1 & 58.1$\pm$5.3 &65.1$\pm$4.2 & 
        
        20.1$\pm$4.1 & 54.7$\pm$4.7 &59.9$\pm$5.8  \\ \bottomrule
        % Add more rows as needed
    \end{tabular}
    \caption{Ratio of successful skill vectors $z$ for each checkpoint (\%). We randomly sampled 100 skills to measure success rate, and repeated ten times to determine the standard deviation.}
    \label{table-success-skill-ratio}
\end{table}

One potential challenge of incorporating a skill discovery module into the learning process is the difficulty of selecting the exact skill that solves the task after training, especially if only a small portion of the skill space is effective. However, we observed that as training progresses, a growing number of skill vectors $z \sim \mathcal{N}(0,I)$ become capable of solving the task. To demonstrate this, we selected model checkpoints at various stages of training and measured the success rate out of random skill vectors. The results are presented in Table \ref{table-success-skill-ratio}. For the leap task, initially, around 43\% of the skills were successful, but this number eventually approached nearly 97\%. Similarly, for the climb and crawl tasks, the proportion of successful skills increased steadily. 

This suggests that once a viable solution is discovered, different skill vectors converge into similar behaviors with the solution. This contrasts with a typical skill discovery scenario where only a small subset of skills solves the task. We observe that this phenomenon of later convergence is facilitated by task rewards: when a skill finds a successful solution, the corresponding trajectory receives higher rewards, which results in the increased probability of the corresponding actions taken. Because all skills share the same policy network, this learning propagates to other skill-conditioned behaviors, leading to what we term a ``positive collapse'' of skills. This is beneficial because it mitigates the issue of selecting the right skill.

\subsection{Wall-jump : Learning Super Agile Tasks}

% \begin{figure}[t]
%     \centering
%     \includegraphics[height=1.1in]{img/wj_cropped2.png}
%     \DeclareGraphicsExtensions.
% \caption{Highly agile behavior discovered by leveraging unsupervised skill discovery as exploration. The quadruped is (1) running toward the wall, (2) jumping off the ground and performing a front flip clockwise, and (3) kicking the perpendicular wall, rotating counterclockwise, and landing.}
% \label{wall}
% \end{figure}

Lastly, we pushed our method to its limits by introducing a new task named \textit{wall-jump}. It requires the robot to perform a sequence of highly agile motions, including running, jumping, flipping, and landing in a specific order. To make this feasible, we devised a guideline-based reward that is widely adopted in robotics\citep{tang2021learning,gu2023rt}. The reward encourages the agent to follow the guideline specified by a user. We used this reward as $r^{\text{task}}$. More details about the reward design can be found in Appendix \ref{appendix_guideline_reward}. The exact guideline is shown in Figure \ref{fig_walljump_solved}(a) in Appendix. Note that the guideline only provides the target trajectory for the root position while not offering any information about orientation.

However, providing the guideline alone was not sufficient for the agent to successfully perform the wall-jump. Figure \ref{fig_walljump_solved}(b) shows the resulting behavior of the agent trained solely with $r^{\text{task}}$. The robot was able to follow the guideline up until it reached the perpendicular wall, but then crashed its back against the wall. The cumulative reward for this episode was about 5.0, as shown by the blue curve in Figure \ref{fig_walljump_solved}(e).  We observe that the robot needs to acquire a specific orientation to kick off the wall and land safely. 

\begin{wrapfigure}{r}{0.3\textwidth}
  \vspace*{-1em}
  \begin{center}
    \includegraphics[width=0.3\textwidth]{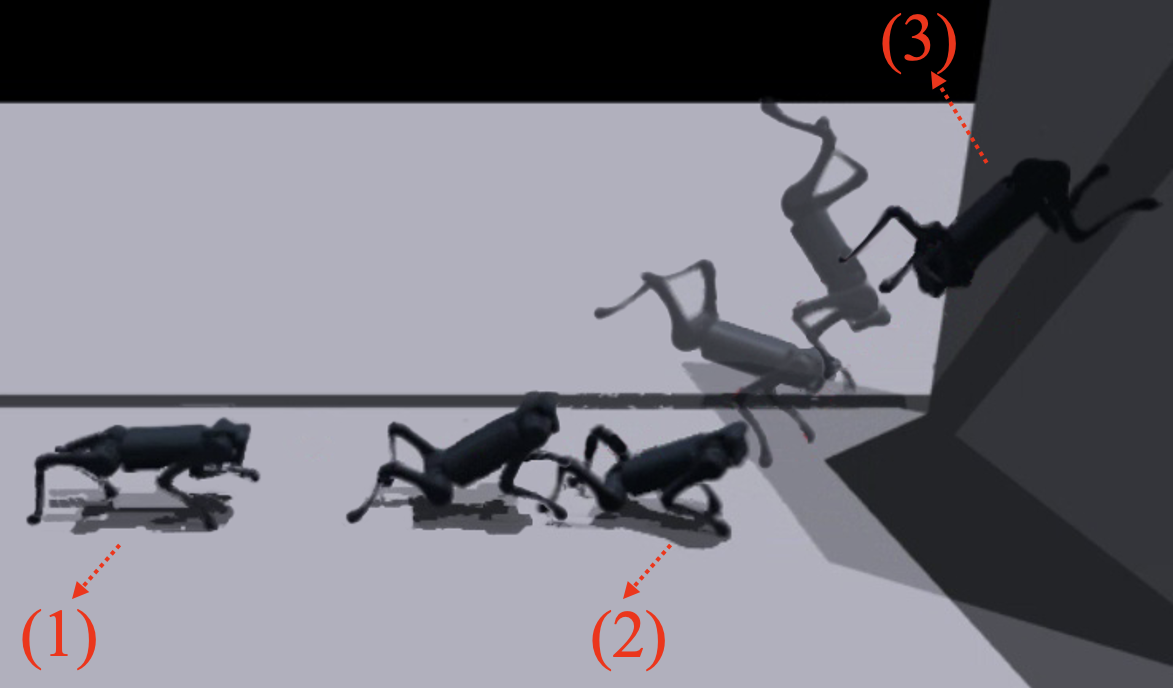}
  \end{center}
  \vspace*{-3.0mm}
  \caption{Wall-jump.}
  \vspace*{-3.0mm}
  \label{fig:walljump}
\end{wrapfigure}
Therefore, we provided the robot's base's {\fontfamily{qcr}\selectfont{\text{roll}}}, {\fontfamily{qcr}\selectfont{\text{pitch}}}, and {\fontfamily{qcr}\selectfont{\text{yaw}}} as input to the skill discovery algorithm, allowing our method to explore and learn diverse orientations of the robot when needed. Figures \ref{fig_walljump_solved}(c) and (d) show the resulting behavior. Our method was able to acquire the specific orientation needed to kick off the wall. As a result, \methodabbr\ achieved a successful wall-jump (Figure~\ref{fig:walljump}) with a much higher task return of 9.5 as indicated by the green curve in Figure \ref{fig_walljump_solved}(e).

\subsection{Hardware Experiments}

% \begin{figure}[t]
%     \centering
%     \includegraphics[height=1.3in]{img/hardware/crawl_carpet.png}
%     \DeclareGraphicsExtensions.
% \caption{Crawling behavior discovered by leveraging unsupervised skill discovery as exploration. }
% \label{fig_hardware_leap}
% \end{figure}

\begin{figure}[t]
    \centering
    {\includegraphics[width=1.0\textwidth]{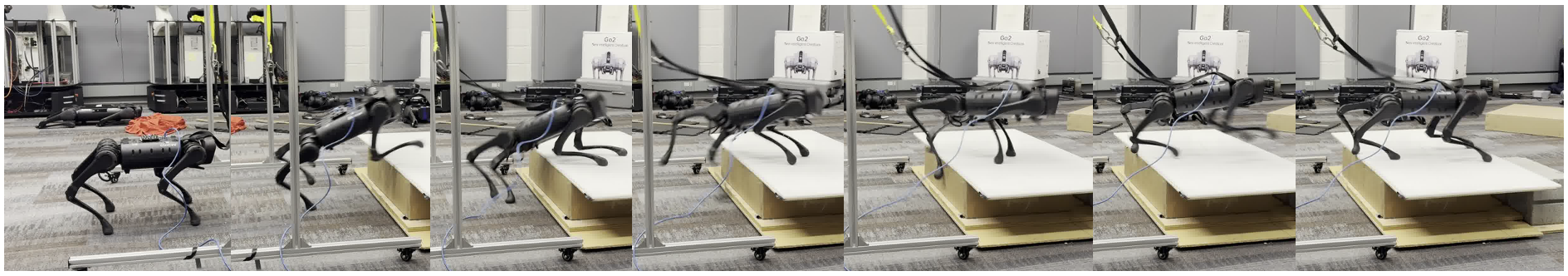}}
    {\includegraphics[width=1.0\textwidth]{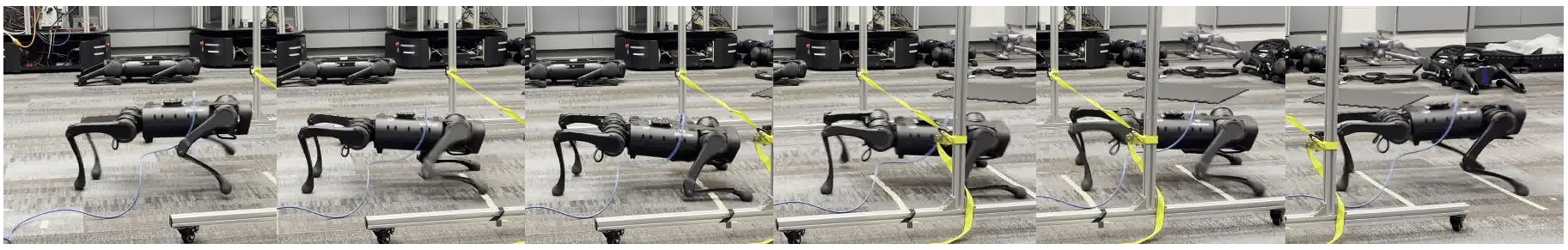}} 
    \DeclareGraphicsExtensions.
    \vspace*{-5mm}
\caption{Both the climbing (top) and crawling (bottom) policies' skills were tested on the real robot.}
\vspace*{-2mm}
\label{fig_hardware_climb_crawl}
\end{figure}

After successfully training a policy that solves the task in simulation, we fine-tune it for real-world deployment by introducing observation noise and domain randomization~\citep{tobin2017domain}. Specifically, we continue training for an additional 5,000 steps with these modifications to improve robustness. Full details of the noise and randomization parameters are provided in Appendix~\ref{observation_noise} and Appendix~\ref{domain_randomization}. To evaluate whether the discovered skills transfer to the real world, we deploy the policy on a Unitree A1 robot. As shown in Figure~\ref{fig_hardware_leap} and Figure~\ref{fig_hardware_climb_crawl}, the robot successfully performs agile maneuvers such as leaping over a $46$~cm gap, climbing a $25$~cm platform, and crawling under a $27$~cm-high obstacle. Furthermore, as illustrated in Figure~\ref{fig_hardware_crawl_diverse_terrain} in Appendix, the crawling policy remains robust even under varying terrain conditions. Please refer to the supplemental video for more details.

%===============================================================================
\section{Conclusion}
In this work, we introduced a novel learning framework that integrates unsupervised skill discovery with reinforcement learning to enable legged robots to acquire highly agile locomotion behaviors without relying on demonstration data or curriculum design. By balancing skill-level exploration and task rewards through a bi-level optimization process, our method allows robots to discover diverse behaviors such as crawling, climbing, leaping, and executing agile maneuvers like wall-jumping. We further demonstrate that the policy learned through our framework can be successfully deployed on real-world hardware.

\section{Limitations}

While our work proposes a novel training framework, it comes with certain limitations. First, effective training requires manual specification of sub-dimensions of the state space to guide exploration. For instance, to induce crawling behavior, we assume that exploring different body heights is essential, and therefore explicitly use height as an input to the diversity objective.

% Another limitation is the two-phase structure of our training pipeline. 
Another empirical observation is that it was better to add observation noise after learning a successful policy using our main algorithm. 
We observed that applying excessive observation noise during skill discovery makes training unstable. This is because the diversity reward relies on distinguishing newly visited states, and noise can obscure meaningful differences, confusing the reward signal. To address this, we first train the skill discovery module under low-noise conditions, and once effective skills are acquired, we fine-tune the policy in a second phase with higher observation noise.  We believe that improving the robustness of the skill discovery module—particularly its ability to operate under noisy observations or heavy domain randomization—would further enhance the applicability and reliability of \methodabbr.

%===============================================================================

%===============================================================================

%===============================================================================

\clearpage
% The acknowledgments are automatically included only in the final and preprint versions of the paper.
\acknowledgments{This research has been funded by the Industrial Technology Innovation Program (P0028404, development of a product level humanoid mobile robot for medical assistance equipped with bidirectional customizable human-robot interaction, autonomous semantic navigation, and dual-arm complex manipulation capabilities using large-scale artificial intelligence models) of the Ministry of Industry, Trade and Energy of Korea.}

%===============================================================================

\bibliography{interplay}  % .bib
\newpage
\appendix

\section{Proof of Equation \ref{equation_last}}
\label{proof_of_bilevel_optimization}

We begin with the Equation (\ref{chain_rule_decomposition}), which is the decomposition of $\nabla_{\lambda}J^{\text{task}}$ using the chain rule,
$$
\nabla_{\lambda}J^{\text{task}} = \nabla_{\theta'}J^{\text{task}} \nabla_{\lambda}{\theta'}.
$$

Here, we can compute the first term $\nabla_{\theta'} J^{\text{task}}$ using the policy gradient theorem~\citep{sutton1999policy}
\begin{equation}
\label{equation_first_term}
\nabla_{\theta'} J^{\text{task}} \approx A^{\text{task}}\nabla_{\theta'} \log \pi_{\theta'}(a | s, z)
\end{equation}

To compute the second term $\nabla_{\lambda} \theta'$, we first derive $\theta'$
\begin{align}
\theta' &= \theta + \alpha \nabla_{\theta} J^{\text{task}+\text{div}}(\theta)  \nonumber \\
&= \theta + \alpha A^{\text{task}+\text{div}} \nabla_{\theta} \log \pi_{\theta}(a | s, z)
\end{align}
Here, $\alpha \in \mathbb{R}$ is a learning rate. Then we can plug in this result to compute $\nabla_{\lambda} \theta'$:
\begin{align}
\label{equation_second_term}
\nabla_{\lambda} \theta' &= \nabla_{\lambda} (\theta + \alpha A^{\text{task+div}} \nabla_{\theta} \log \pi_{\theta}(a | s, z)) \nonumber \\
&= \nabla_{\lambda} (\alpha A^{\text{task+div}} \nabla_{\theta} \log \pi_{\theta}(a | s, z)) \nonumber \\
&= \nabla_{\lambda} (\alpha A^{\text{task}} + \alpha \lambda A^{\text{div}})  \nabla_{\theta} \log \pi_{\theta}(a | s, z) \nonumber \\
&= \alpha A^{\text{div}} \nabla_{\theta} \log \pi_{\theta}(a | s, z)
\end{align}

Finally, we can compute the value of $\nabla_{\lambda}J^{\text{task}}$ by plugging in the Eq. (\ref{equation_first_term}) and Eq. (\ref{equation_second_term}):  
\begin{equation}
\nabla_{\lambda}J^{\text{task}} \approx 
A^{\text{task}} \nabla_{\theta'} \log \pi_{\theta'}(a | s, z) *  \alpha A^{\text{div}} \nabla_{\theta} \log \pi_{\theta}(a | s, z)
\end{equation}

This concludes the derivation of the equation \ref{equation_last}.

\section{Crawl on Diverse Terrain}

\begin{figure}[h]
    \centering
    % \subfigure[]
    {\includegraphics[width=1.0\textwidth]{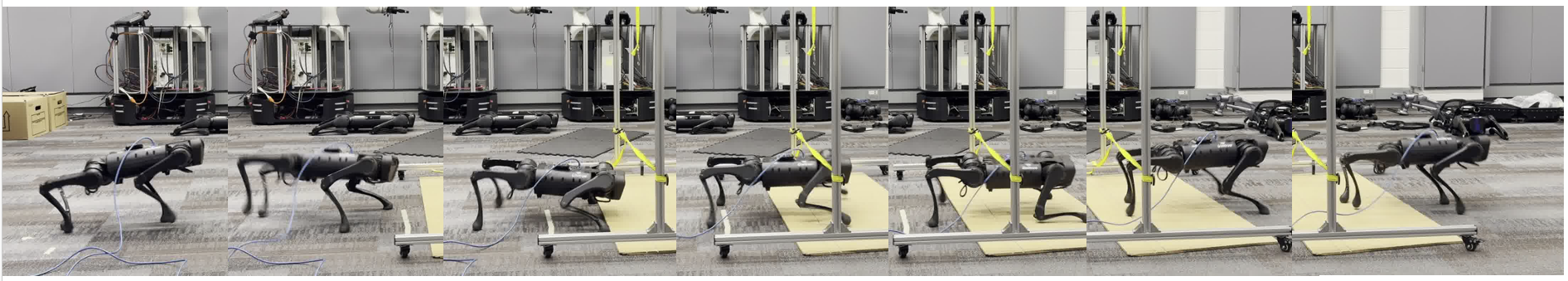}}
    % \hspace*{-2mm}
    % \subfigure[]
    {\includegraphics[width=1.0\textwidth]{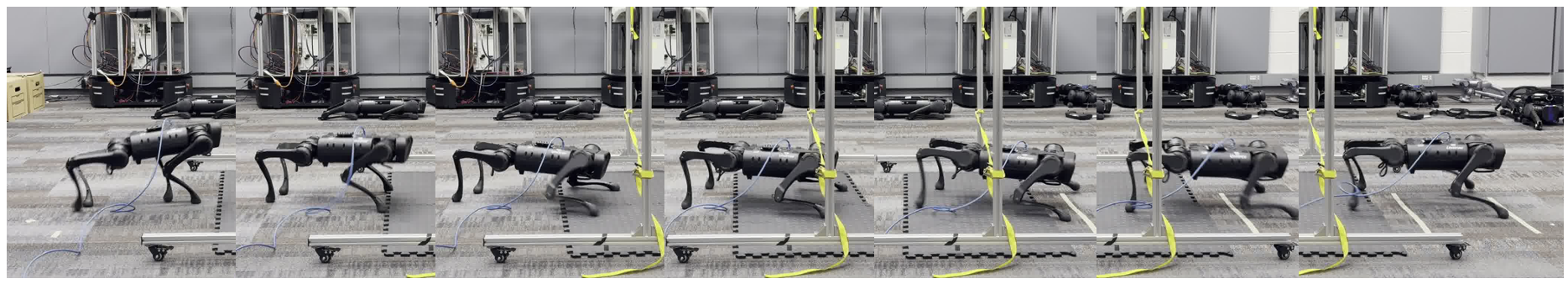}} 
    \DeclareGraphicsExtensions.
\caption{The crawling policy demonstrates robustness by successfully navigating under obstacles on different terrains: wood (top) and rubber mat (bottom).}
\label{fig_hardware_crawl_diverse_terrain}
\end{figure}

To evaluate the robustness of the crawling policy trained with our skill discovery approach, we deployed it on two different terrains: wood and a rubber mat. As shown in Figure~\ref{fig_hardware_crawl_diverse_terrain}, the robot successfully crawled under the obstacle in both settings, demonstrating its ability to generalize across varying surface conditions.

\newpage

\section{Implementation Detail}
\label{appendix_implementation_detail}

\subsection{Algorithm}
\begin{algorithm}[h]
\label{algorithm_ours}
\caption{}
\begin{algorithmic}[1]
\STATE Initialize skill-conditioned policy $\pi_{\theta}(a|s, z)$, value functions $v^{\text{task}}_{\psi_1}$ and $v^{\text{div}}_{\psi_2}$, representation function $\phi(s)$, Lagrange multiplier $\kappa$, Balancing parameter $\lambda$, data buffer $\mathcal{D}$
\FOR{$i \leftarrow 1$ to \# of epochs}
    \FOR{$j \leftarrow 1$ to \# of episodes per epoch}
        \STATE Sample skill $z \sim \mathcal{N}(0, I)$
        \WHILE{episode not terminates}
            \STATE Sample action $a \sim \pi(a|s, z)$ 
            \STATE Execute $a$ and receive $s'$ and $r^{\text{task}}$
            \STATE Compute $r^{\text{div}} =(\phi(s') - \phi(s))^T z $
            \STATE Add $\lbrace s, a, r^{\text{task}}, r^{\text{div}}, s' \rbrace$ to data buffer $\mathcal{D}$
        \ENDWHILE
    \ENDFOR
    \FOR{ $\lbrace s, a, r^{\text{task}}, r^{\text{div}}, s' \rbrace$ in $\mathcal{D}$}
        \STATE Update $\phi(s)$ to maximize $\mathbb{E}_{(s,z,s')\sim\mathcal{D}}\left[ (\phi(s') - \phi(s))^T z + \kappa \cdot \min(\epsilon, 1 - \lVert \phi(s) - \phi(s')\rVert_2^2) \right]$
        \STATE Update $\kappa$ to minimize $\mathbb{E}_{(s,z,s')\sim\mathcal{D}}\left[ \kappa \cdot \min(\epsilon, 1 - \lVert \phi(s) - \phi(s')\rVert_2^2) \right]$
        \STATE Update $\theta$ using PPO with reward $r = r^{\text{task}} + \lambda * r^{\text{div}}$
        \STATE Update $\psi_1$ and $\psi_2$ using $r^{\text{task}}$ and $r^{\text{div}}$ respectively
        \STATE Update $\lambda$ using Eq. \ref{equation_last}
    \ENDFOR
    
\ENDFOR
\end{algorithmic}
\end{algorithm}

\subsection{Task reward detail}
\label{appendix_reward}

\begin{table}[h]
    \centering
  \caption{Task rewards}
  \begin{tabular}{lcr}
    \toprule
    Name     &  Mathematical Expression& Coefficients value \\
    \midrule
    Tracking angular velocity & $e^{-|w_{yaw}|}$ &0.05\\
    Tracking linear velocity & $|v_x - v_x^{target}| $ & -1\\
    Alive & - &2 \\
     Torque squared & $\sum\limits_{j \in \text{joints}} |\tau_j \dot{q}_j| ^2 $ &-1e-6\\
    Exceed dof pos limits  &$\sum\limits_{j \in \text{joints}} \max(|\text{dof}_j| - \text{dof}_{\text{lim}}, 0)$ &-0.1 \\
    Exceed torque limits & $\sum\limits_{j \in \text{joints}} \max(|\tau_j| - \tau_{\text{lim}}, 0)$ &-0.2 \\

    \bottomrule 
  \end{tabular}
  
  \centering
  
\end{table}

The first three terms about tracking commands specify the goal of the task, while the other three terms regularize unrealistic, infeasible motions.

\newpage

\subsection{Observation space}
\label{appendix_observation}

\begin{table}[h]
    \centering
  \caption{A1 Robot Observations}
  \begin{tabular}{lll}
    \toprule
    Name     & Description     & Dimension \\
    \midrule
    Base position & x,y,z position of the robot's base  & 3     \\
    Base rotation & Yaw, Pitch, Roll of robot's base & 3\\
    Base velocity & Velocity of robot's base in x,y,z direction & 3 \\ 
    Base ang vel & Angular velocity of robot's base & 3 \\
     
    Gravity projection & Vector indicates direction of the gravity & 3\\ 
    Velocity command     & Velocity command given by users & 3      \\
    DOF position     & Current angle of each DOF & 12      \\
    DOF velocity     & Angular velocity of each DOF      & 12  \\
    Previous action     & Action executed in previous step     & 12  \\
    Distance to obstacle    & Distance to obstacle     & 1  \\
    Obstacle properties & Difficulty and length of the obstacle & 2 \\
    Obstacle info & One hot encoding to identify the obstacle type& 5\\
    Sidewall distance    & Distance to side wall     & 2  \\
    Sampled skill   & Sampled skill for current episode     & 1  \\
   
    \midrule
    Sum & &  65 \\
    \bottomrule 
  \end{tabular}
  
  \centering
  
\end{table}

In the hardware experiments, we used a motion capture system to obtain global measurements such as the robot’s base position and velocity. We also experimented with including certain robot configuration parameters—such as mass and motor strength—as part of the observation. Since these parameters are used in domain randomization, providing them could help the policy to be aware of the current configuration. However, given that training was successful both with and without these configuration inputs in both simulation and real-world settings, we conclude that this information is not critical for policy performance.

\subsection{Observation noise}

\label{observation_noise}
\begin{table}[h]
    \centering
    \caption{Observation noise coefficients per task}
    \begin{tabular}{lccc}
        \toprule
        Obs & Crawl & Leap & Jump \\
        \midrule
        Base position & $\pm$0.1 & $\pm$0.1 & $\pm$0.1 \\
        Base rotation & $\pm$0.1 & $\pm$0.1 & $\pm$0.1 \\
        Base lin vel & $\pm$0.1 & $\pm$0.1 & $\pm$0.1 \\
        Base ang vel  & $\pm$0.2 & $\pm$0.2 & $\pm$0.2 \\
        DOF position & $\pm$0.01 & $\pm$0.01 & $\pm$0.01 \\
        DOF velocity  & $\pm$1.5 & $\pm$1.5 & $\pm$1.5 \\
        Gravity projection  & $\pm$0.05 & $\pm$0.05 & $\pm$0.05 \\
        Distance to obstacle & $\pm$0.0 & $\pm$0.05 & $\pm$0.05 \\
        \bottomrule
    \end{tabular}
\end{table}

\subsection{Domain randomization parameters}

\label{domain_randomization}
\begin{table}[h]
    \centering
    \caption{Domain Randomization Parameters}
    \begin{tabular}{lcc}
        \toprule
        Parameter & Range & Form \\
        \midrule
        Mass & [-1.0, 3.0]  & Additive \\
        Friction & [0.0, 2.0] & Multiplicative \\
        Motor strength & [0.9, 1.1] & Multiplicative \\
        Init base x position & [0.2, 0.6] & Additive \\
        Init base y position & [-0.25, 0.25] & Additive \\
        Init DOF position range & [0.5, 1.5]& Multiplicative \\
        \bottomrule
    \end{tabular}
\end{table}

Domain randomization parameters were sampled from a uniform distribution with the range above.

\subsection{Hyperparameters}
\label{appendix_hyperparam}

\begin{table}[h]
  \caption{Hyperparameters of our method}
  \centering
  \begin{tabular}{lr}
    \toprule
    Name     & Value         \\
    \midrule
    Learning rate &  0.0005    \\
    Optimizer     &  Adam\citep{kingma2014adam}   \\
    PPO clip threshold  &  0.2   \\
    PPO number of epochs  & 5  \\
    GAE $\lambda$ \citep{schulman2015high}  &  0.95  \\
    Discount factor $\gamma$  & 0.99 \\ 
    Horizon length   & 24 \\
    Entropy coefficient  &   0.001  \\
    
    Policy network $\pi$ &  MLP with [512, 256, 128], \\
    Activation of $\pi$&  ELU\citep{clevert2015fast} \\
    Value network $v$  & MLP with [512, 256, 128] \\
    Activation of $v$ &  ELU\citep{clevert2015fast} \\
    Representation function $\phi$ from METRA &  MLP with 
        [256, 256, 256]\\ 
    Activation of $\phi$ &  ReLU \\
    Initial Lagrange coefficient $\kappa$ from METRA&  30 \\
    \bottomrule
  \end{tabular}
\end{table}

\newpage

\section{Details of the Wall-jump experiments}

\begin{figure*}[h]
    \centering
    \subfigure[Red dots - Human drawn guidelines]
    {\includegraphics[width=0.45\textwidth]{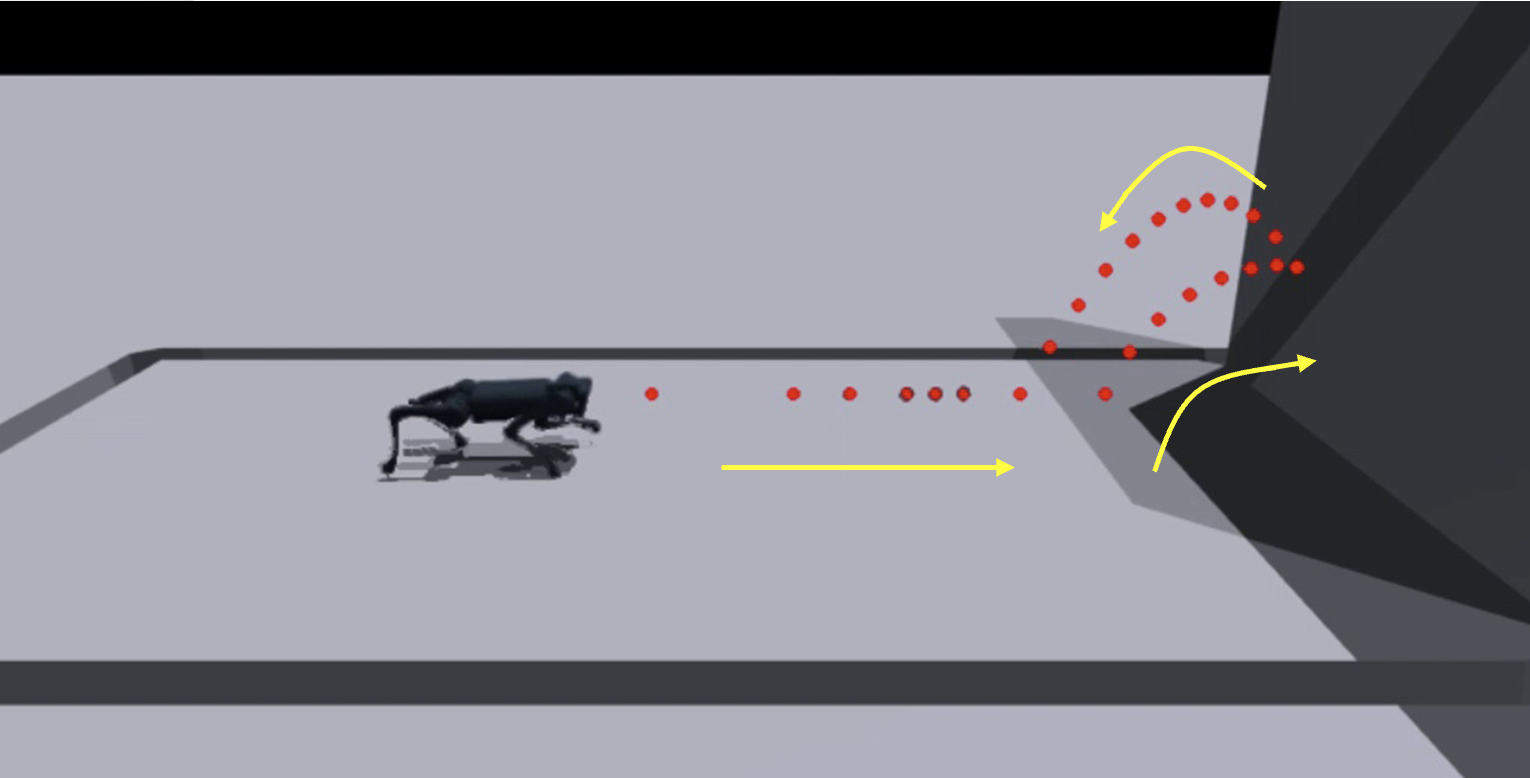}} 
    \subfigure[\textit{Task-only} makes robots crash]
    {\includegraphics[width=0.45\textwidth]{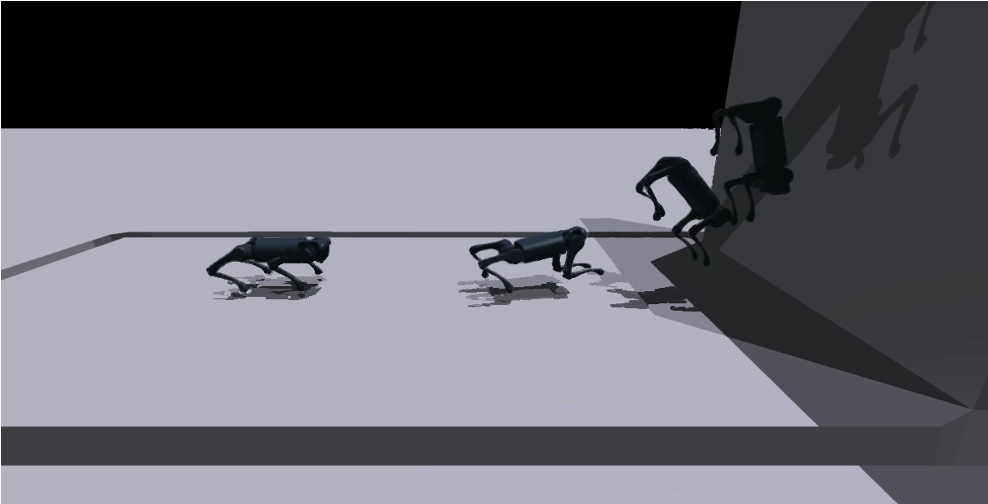}} 
    \vspace*{-2.0mm}
    
    \subfigure[Ours- A robot runs and front-flips to kick wall]
    {\includegraphics[width=0.45\textwidth]{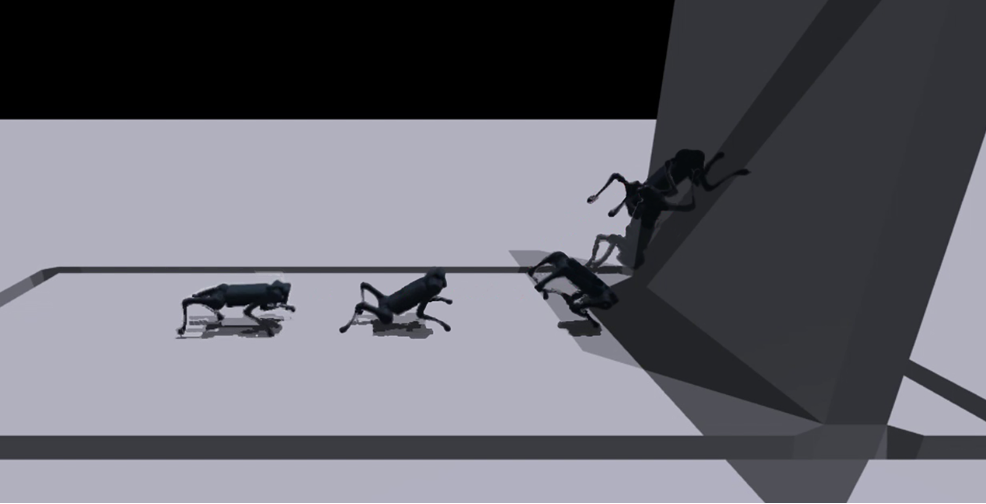}} 
    \subfigure[Ours- using wall, performs back-flip and lands]
    {\includegraphics[width=0.45\textwidth]{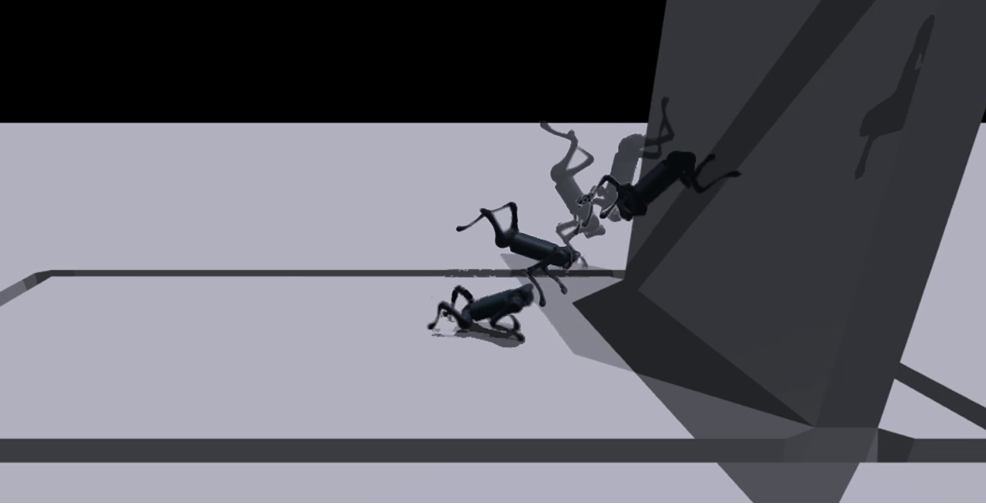}} 

    \vspace*{-4.0mm}
    \subfigure[Training curves]
    {\includegraphics[width=0.48\textwidth]{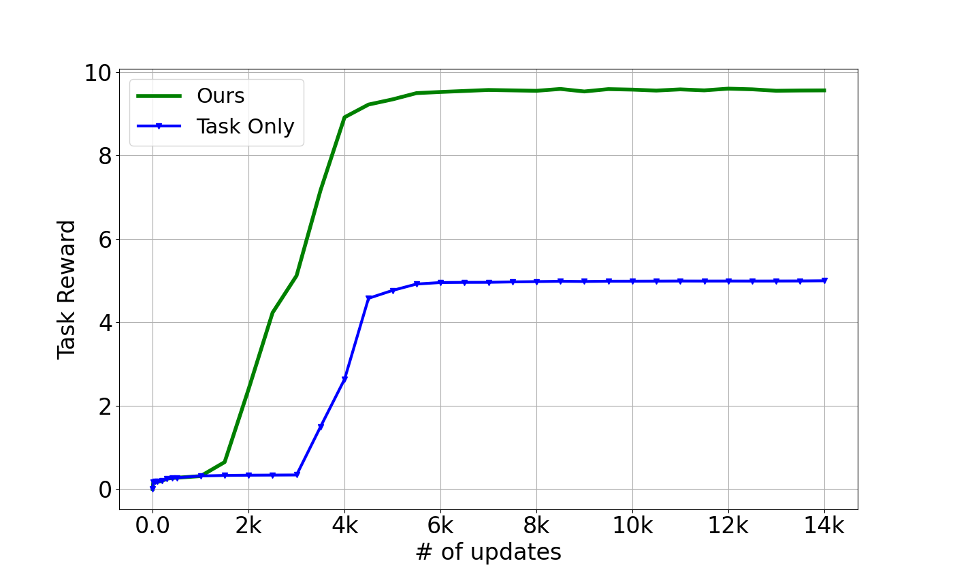}}
    \subfigure[$\lambda$ curve]
    {\includegraphics[width=0.48\textwidth]{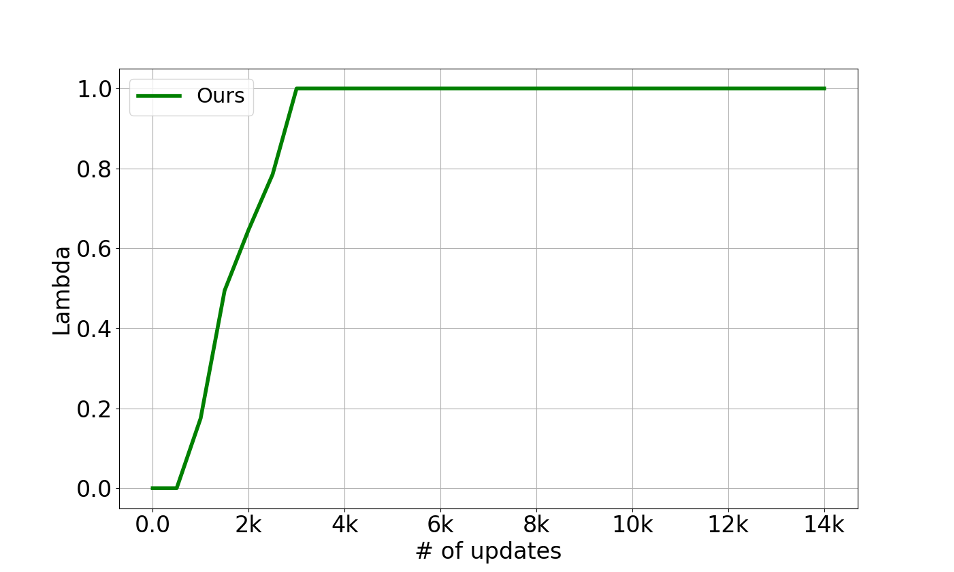}} 
    \caption{Our method enables robots to solve the wall-jump task.}
    \label{fig_walljump_solved}
\end{figure*}

\subsection{Details of the guideline following reward}
\label{appendix_guideline_reward}
For the wall-jump task, we defined a special task reward, $r^{\text{task}}$, based on a guideline provided by a human. The guideline consists of a sequence of $n$ points:
$$g_{i=0,1,...,n-1} \in \mathbb{R}^3$$
Let the robot's base position in global 3D space be denoted as $\bm{x} \in \mathbb{R}^3$. At each time step, the robot has a target point $g_i$, starting with $g_0$. When the robot reaches the current target, it moves on to the next target, $g_{i+1}$. A target is considered reached when the distance between $\bm{x}$ and $g_i$ falls below a threshold $h \in \mathbb{R}$, i.e., $||\bm{x} - g_i||_2 < h$.

Then, the reward can be defined as follows:
$$ r_t = e^{-||\bm{x} - g_i||_2}$$
This term has the desirable property of being bounded between 0 and 1. It approaches 0 when the robot is infinitely far from the current target and becomes 1 when the robot exactly reaches the target. This property contributes to stability during the learning process. We optimized this reward using reinforcement learning (RL) to train the agent to follow the given guideline.

% \section{Additional Task Related Plots}
% \label{appendix_torue_plots}
% \def\scaleQS{1.1}

% \begin{figure*}[!h]
%     \centering
%     \subfigure[Leap]
%     {\includegraphics[width=1\textwidth]{ICLR 2025 Template/leap_torques.png}}
%     \hspace*{-5mm}
%     \subfigure[Climb]
%     {\includegraphics[width=1\textwidth]{ICLR 2025 Template/jump_torques.png}}
%     \hspace*{-5mm}
%     \end{figure*}
%     \begin{figure*}[!h]
%     \centering
%     \ContinuedFloat
%     \subfigure[Crawl]
%     {\includegraphics[width=1\textwidth]{ICLR 2025 Template/jump_torques.png}}
    
%     % \subfigure[Crawl]
%     % {\includegraphics[width=1.1/textwidth]{ICLR 2025 Template/crawl_torques.png}} 
%     \caption{Torque Plots for different tasks }
%     \label{fig_torque}
% \end{figure*}

\end{document}